# A Survey of Recent Developments in Collision Avoidance, Collision Warning and Inter-Vehicle Communication Systems

Öncü Ararat and Bilin Aksun Güvenç

Department of Mechanical Engineering, Istanbul Technical University,

Gümüşsuyu, Taksim, İstanbul, TR-34437, Turkey

*Abstract*—This paper presents the state-of-the-art on Collision Avoidance/Collision Warning (CA/CW) systems. Traffic accidents result from driver errors or situations that are unpredictable for the driver. CA/CW systems are developed for reducing traffic accidents and saving peoples' lives by warning drivers or taking compensatory action when drivers can not react fast enough. Besides these initiatives, this paper explains the importance of CA/CW driving assistance systems by considering economic issues as well. Technological developments are investigated in the context of finished and ongoing projects all over the world. CA/CW system algorithms are discussed regarding performance criteria such as reliability and strictness. This paper also presents the information on Inter-Vehicle Communication Systems (IVC) which will be a key ingredient of future CA/CW systems.

## I. INTRODUCTION

Traffic crashes cause deaths of millions of people every year. Researchers tried to overcome this fatal problem by developing passive safety systems like seat belts, air bags and crash zones [1]. Although these passive measures helped a lot; there must be more effective ways of holding accidents at acceptable levels. This goal is closer to realization through advances in preventive and active safety systems.

Despite the fact that research on these subjects began with Autonomous Highway Systems (AHS), the auto industry has turned to Advanced Driving Assistance Systems (ADAS) considering the reality that progressions on AHS can only be realized in the long term period [2].

Another reason for the great interest on ADAS is economic. Driving assistance systems increase the efficiency of traffic and reduce fuel consumption [3]. Highway capacity is related to both vehicle speed and inter-vehicle spacing [4]. With more intelligent vehicles, it is possible to realize fast vehicles following each other with small inter-vehicle spaces safely and this leads to increased highway capacity. On the other hand, low velocity changes and the ability of vehicles to communicate with each other will diminish fuel consumption [3].

CA/CW systems are an important part of ADAS which are composed of systems such as lane keeping, lane departure prevention, driver condition monitoring, night vision, adaptive cruise control and stop and go assistants. USA, Japan and the European Union have a leading role in research on CA/CW systems.

This paper presents the state-of-the-art on CA/CW systems and is organized as follows: In section II, recent developments and ongoing projects on CA/CW systems in USA, Japan and the European Union are presented. In section III, algorithms developed in these projects are listed. In section IV, Inter-Vehicle Communication Systems and the protocols used in these systems are described. The paper ends with conclusions in section V.

## II. CA/CW SYSTEMS

Collision Avoidance/Collision Warning systems have received great attention all over the world. Despite this attention, countries are on different positions on the progression of this technology. USA, Japan and Europe have a leading role on CA/CW research.

Research in USA on CA/CW systems is executed mainly by the Departments of Transport (DOT) at state and national level despite lack of hierarchy between them. Beginning with AHS studies in Federal Highway Administration (FHWA), research turned to ADAS in National Highway Traffic Safety Administration-Office of Crash Avoidance Research (NHTSA-OCAR) at the national level. The state level research was started by California DOT. Minnesota, Virginia and Arizona DOT followed California [2].

NHTSA-OCAR keeps on CA/CW systems research within the Intelligent Vehicle Initiative (IVI) project. Organizations which are composed of universities, industrial and public foundations are responsible for specific subjects of these research efforts. To obtain effective systems, human factors and legal issues were investigated after determining the most common crash types and searching for their causes. Results of studies on forward collision avoidance, rear-end-collision avoidance and intersection collision avoidance systems were tested in Field Operational Tests (FOTs) which were also part of the IVI project [5].

Since its main aim was to develop a reliable system which had market potential and easily adaptable characteristics, an extensive part of the forward collision avoidance system project coordinated by NHTSA-OCAR was allocated to cheap and reliable component development. Cheaper materials for Microwave Monolithic Integrated Circuit (MMIC) and Transferred-Electron (GUNN) receivers were tested. Another related research was carried out for developing cheap laser sensors. The component development phase was followed by an algorithm development phase. The developed system determined the radius of curvature by using yaw, speed and steering sensors. By determining the radius of curvature, it was possible to determine the presence of vehicles traveling on the host vehicle's lane with the aids of radar data. Then relative parameters were calculated and compared with the critical parameter values [1].

The NHTSA-OCAR study on rear-end-collision avoidance system was at the algorithm development level. In this research, kinematic analyses of the selected scenarios were carried out to build an overall algorithm. Critical warning distance was determined according to the appropriate scenario which best represented the present status of the two vehicles [6].

Intersection collision avoidance systems were studied by two groups in NHTSA-OCAR. The first project aimed at overcoming specific intersection collision scenarios. The developed system used both Differential Global Positioning System (DGPS) and Geographical Information System (GIS). The system took vehicle position, velocity and heading information from DGPS. GIS gave intersection point and roadway geometry information. GIS also gave traffic control infrastructure information. If traffic control unit sign and sign phase time allowed the vehicle to pass through the intersection, then whether the vehicle could stop with a maximum deceleration of 0.35g or not was controlled. If it could, then whether there was any collision risk or not with the crossover vehicle was controlled. If there was no risk, the system let the vehicle pass through the intersection. Otherwise, the driver was warned by the system [7].

The second project in intersection collision avoidance aimed at developing a system that worked regardless of the direction of the vehicles. This system did not need any GIS information. All of the information was obtained by inter-vehicle sensors and Inter-Vehicle Communication Systems. Relative position of the vehicles was calculated according to a fixed coordinate system [8].

California DOT carries out research on CA/CW systems via its Partners for Advanced Transit and Highways (PATH) program. A detailed rear-end collision avoidance system was created within that project.

In another research effort, a nonlinear controller for this rear-end collision avoidance system was developed by modeling the brake and the vehicle as a whole. The brake system was modeled as second order differential equation. The maximum jerk was held between -10 and 10 m/s3, the desired acceleration/deceleration profile was selected as a sine function within the start and stop periods. After inserting brake pressure into the vehicle model, a sliding surface for brake pressure was defined and a sliding mode controller was applied [10].

Simulation of this system was done using Hardware-In-the-Loop (HIL) testing. The controller of this system was installed on a DSP chip. Velocity and position information of the leading and following vehicles were taken from a vehicle model. With the calculated values, a DC motor which was installed as hardware controlled the relative distance and velocity. Measured values of relative distance and velocity taken from a radar sensor were fed into the controller [11].

Japan is one of the big actors on ADAS research. Research in Japan is coordinated by the government via three ministries: Ministry of Construction (MOC), Transport (MOT) and International Trade and Industry (MITI) [2].

MOC supports ADAS research by means of Advanced Cruise-Assist Highway Systems Research Association (AHSRA). AHSRA was set up after two successful projects on ADAS. This association carries out the infrastructure research in Japan. Initial projects were on the development of system components. In the second stage, required services and essential studies for these services were determined. Standards, bases and education for popularization were created for developed projects to provide widespread usage. Last stage composed of verification and demonstration tests of developed systems. The next objective is converting the developed systems to viable and practical applications [34].

Intelligent vehicle research is carried out by Advanced Safety Vehicle (ASV) project which is sponsored by MOT. Success of the first part which was finished in 1996 provided powerful support for the second part. Unlike AHSRA, research in ASV was carried out by industrial organizations individually rather than collective studies. Accident avoidance system for intersection, collision avoidance system by means of inter-vehicle communication and collision avoidance system by means of road-vehicle communication are ongoing projects in ASV. Activities for understanding ASV technologies, studies on telecommunication standards for ASV and standardization of ASV technologies are further issues which are intended to be realized in the short term period [35].

MITI has a smaller workspace on ADAS than the other two ministries. Research on assistance systems was controlled by its Mechanical Engineering Laboratory which is sponsored by the ministry. Research concentrated on ITS and Platoon Systems. The Super-Smart Vehicle System (SSVS) research project dealt with a wide range of topics on ADAS and was coordinated by this laboratory. These systems were based on inter-vehicle communication and GPS [2].

Industrial research organizations had their own projects on ADAS along with supporting projects which were coordinated by the government. Honda and Mazda, for example, developed a kinematic analysis based rear-end collision avoidance algorithm.

Research in Europe on ADAS kept on with great success resulting from very well planned projects. There were three ways of research on ADAS in Europe: industrial research, governmental research projects and European Union projects.

Governments were influential on solving arising traffic crash problems which caused deaths of millions of people and enormous economic losses each year. France was one of these countries interested in conquering this problem as it had the largest transportation network in Europe and accidents resulting in deaths of 8000 people per year. Research in France was coordinated by four laboratories which were The National Research Laboratory on Transport and Safety (INRETS), The National Research Laboratory on Road Infrastructures (LCPC), The National Research Laboratory on Informatics and Automation (INRIA) and The National Scientific Research

Organization (CNRS). INRETS and LCPC opened a joint research laboratory which was called LIVIC. This laboratory had projects on interactions between road, vehicle and driver. There were also demonstration programs that tested the developed technologies. Another organization studying this subject was LARA which was created by INRETS, LCPC, INRIA, CNRS and top universities of the country. In addition to ADAS projects, there were also research efforts on AHS [2].

Holland is another important country in Europe in ADAS.Rijkswaterstaat (RWS), which is a Directorate of the Ministry of Transport and Public Works in the Holland, had started ADAS and AHS project cooperation with the Netherlands Organization for Applied Scientific Research (TNO) [2].

Simulation results were then controlled in a Hardware-in-the-Loop (HIL) system called VEHIL. VEHIL contained two HIL systems. The first one was used for longitudinal technology testing. The host vehicle was called Vehicle under Test (VUT) in this system and it had headway sensors and receivers as well as other vehicles which were on the platform simulating the real road in an environment animated by a robot vehicle. Lateral tests were realized using Automatic Guided Vehicle (AGV) which can move in the lateral and longitudinal directions. Underground Logistical System (OLS) vehicle was positioned on the AGV. AGV determined its position by inter-vehicle sensors where OLS determined its position with the magnetic grids. Magnetic grids were formed by Grid Simulator which was composed of three metal shaft and three magnetic disks rotating around this shaft. In this way, simulation of OLS which would be tested in real magnetic environment was realized [15].

European Union's Framework Programs had been constructed for creating a centralized research organization studying new technologies and their social and legal results. 6th Framework Program which is the latest and continuing one was composed of four areas which are thematic research, strengthening the foundations of ERA, structuring ERA and nuclear energy. There are also cross-cutting research activities consisting of multidisciplinary research. ADAS research is realized in thematic and cross-cutting research activities [36].

Prometheus was the first project on ADAS realized by the EU Union which started in 1986 and finished with the VITA II demonstration in 1994. Studies on AHS and ADAS were directed by Daimler-Benz. Lane departure warning, lane obstacle warning and adaptive cruise control were three topics of the project which was categorized as a Europe-wide Network for Industrial R&D (EUREKA) project [2].

Mobility and Transport in Intermodal Traffic (MOTIV) was a regional research project supported by the EU Union. This project was composed of two parts which were Mobility in Urban Areas and Safe Roads. Safe Roads was a kind of ADAS project which consisted of turn-off and lane-change assistance, adaptive cruise control, vehicle-vehicle communication, driver assistance strategies and man-machine interaction topics [2].

PREVENT is composed of five parts which are: safe speed and safe following, lateral support, intersection safety and vulnerable road users. Horizontal activities consist of three component improvement projects: RESPONSE 3, MAP&ADAS and Profusion. RESPONSE 3 investigates the legal issues and human factors. Digital maps for ADAS are being studied in MAP&ADAS. Profusion is a sensor improvement project. Safe speed and safe following are projects on longitudinal collision avoidance systems and contain two subprojects: SASPENSE and WILLWARN. SASPENSE aims at avoiding collisions with road condition, traffic density, road geometry, frontal obstacles, potentially dangerous road locations and weather condition information by using vehicular sensors. WILLWARN uses radars as inter-vehicle communication system and GPS data to prevent possible accidents. SAFELANE and LATERAL SAFE form the third part of the project called lateral support. Purpose of SAFELANE is developing a lane departure warning system which uses radar, camera and digital map information. Lane changing assistant system that detects side obstacles and warns driver of blind zone, dangerous lane changing and merging maneuvers is a subject of LATERAL SAFE. Failing to distinguish other drivers' actions, missing road signs or inappropriate maneuvers are the most common sources of intersection collisions. The main aim of INTERSAFE was to remove these sources by fusion of radar-camera sensors and information from traffic signals. Last part of PREVENT is formed for collision mitigation for crashes including vulnerable road users. It consists of two subprograms APALACI and COMPOSE, which use different algorithms for sensor fusion to detect pedestrians and motorcycle drivers [16].

III. CA/CW SYSTEMS ALGORITHMS

In this section rear-end and intersection collision avoidance system algorithms are presented.

*A. Rear-End Collision Avoidance Systems Algorithms*

Rear-end collision avoidance system algorithms are based on the tendency of a rear-end collision by calculating critical distances between two vehicles following each other and comparing this with present distances to apply appropriate actions for different scenarios. The basic geometry of a rear-end collision is given in Figure 1.

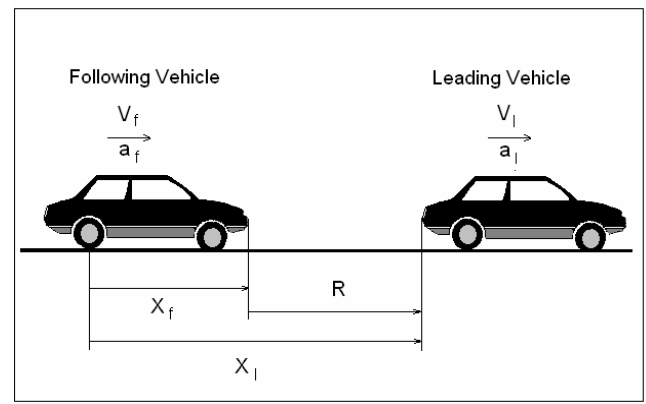

Figure 1. Basic Geometry of Rear-End Collision Avoidance

The rear-end collision avoidance system algorithm developed by NHTSA-OCAR was an overall algorithm which was built for a situation of two vehicles following each other with constant speed while the leading vehicle began to brake. Three scenarios were created; leading vehicle stopping before collision warning alarm, leading vehicle stopping after collision alarm before following vehicle stopped and leading vehicle stopping after following vehicle stopped. Boundary analyses of these three scenarios were done in the time domain initially. Boundary for scenario 1 and 2 was stopping leading vehicle at the time of warning and boundary for scenario 2 and 3 was stopping leading vehicle and following vehicle at the same time.

Boundary value formulation for scenario 1 and 2 was;

$$T_h = \frac{1}{2}v_0(\frac{1}{a_l} + \frac{1}{a_f}) + \frac{6.67}{v_0} + 1.5 \quad (1)$$

Boundary value formulation for scenario 2 and 3 was;

$$T_h = \frac{1}{2}v_0(\frac{1}{a_l} - \frac{1}{a_f}) + \frac{6.67}{v_0} \quad (2)$$

where $a_l$ and $a_f$ are decelerations of leading and following vehicles respectively. $v_0$ is the initial velocity, $R_0$ is initial distance between the two vehicles, $T_h$ is headway, $R_0 / v_0$ is initial headway, 1.5 seconds is driver delay and 6.67 ft (2 m) is safe stopping distance. With the initial velocity and the deceleration of the leading vehicle it was possible to calculate headway value. Initial headway value was compared with headway values from (1) and (2) and the present states for the following vehicle to determine the appropriate scenario. Then present distance between the two vehicles was compared with the critical warning distance for the appropriate scenario which was calculated using equations (3), (4), (5), (6) and (7).

First scenario;

$$R_W = \frac{1}{2}\frac{v_0^2}{a_f} + 1.5v_0 + 6.67 \quad (3)$$

Second scenario;

$$t_w = \frac{1}{2}v_0(\frac{1}{a_l} - \frac{1}{a_f}) + (T_h - 1.5) - \frac{6.67}{v_0} \quad (4)$$

$$R_W = R_0 - \frac{1}{2}a_l t_w^2 \quad (5)$$

Third scenario;

$$t_w = \left[\frac{(a_f - a_l)}{a_f}\right]\left[2\frac{(v_0 T_h - 6.67)}{a_l(1 - \frac{1}{a_f})}\right]^{1/2} - 1.5 \quad (6)$$

$$R_W = R_0 - \frac{1}{2}a_l t_w^2 \quad (7)$$

where $R_W$ is warning distance and $t_w$ is warning time [6]. Another Rear-End Collision Avoidance System was developed in California PATH Program. This system used a non-dimensional parameter for determining warning and braking time as

$$w = \frac{d - d_{br}}{d_w - d_{br}} \quad (8)$$

$$d_w = \frac{1}{2}\left(\frac{v_f^2}{a_f} - \frac{(v_f - v_{rel})^2}{\alpha_l}\right) + v_f\tau + d_0 \quad (9)$$

$$d_{br} = v_{rel}(\tau_{sys} + \tau_{hum}) + 0.5a_f(\tau_{sys} + \tau_{hum})^2 \quad (10)$$

where $a_l$ and $a_f$ are the decelerations of the leading and following vehicles, respectively. d is the present distance between two vehicles, $d_{br}$ is the braking distance, $d_w$ is the warning distance, $\tau$ is the total delay, $\tau_{sys}$ is the system delay, $\tau_{hum}$ is the driver delay, $v_f$ is the following vehicle velocity, $v_{rel}$ is the relative velocity and w is the dimensionless parameter. w>1 describes safety zone, 0<w<1 describes alarm zone where being closer to zero results in stronger alarms. w<0 describes the braking zone. To obtain more effective systems, these distances were corrected by road and human factors as

$$d_{w,scaled} = d_w f(\mu)g(driver) \quad (11)$$

$$d_{br,scaled} = d_{br} f(\mu)g(driver) \quad (12)$$

where $d_{w,scaled}$ and $d_{br,scaled}$ are the corrected critical warning and braking distances, respectively. $f(\mu)$ is the road friction factor function and g(driver) is the human factor function [9].

Mazda used a kinematic analysis to determine the critical braking distance on its rear-end braking system algorithm using

$$d_{br} = \frac{1}{2}\left[\frac{v_f^2}{a_f} - \frac{(v_f - v_{rel})^2}{a_l}\right] + v_f\tau_1 + v_{rel}\tau_2 + d_0 \quad (13)$$

where $a_l$ and $a_f$ are the maximum decelerations of the leading and following vehicles, respectively. $d_{br}$ is the braking distance, $v_f$ is the following vehicle velocity, $v_{rel}$ is the relative velocity, $\tau_1$ is the driver delay, $\tau_2$ is the brake system delay, $d_0$ is the distance between the two vehicle [12].

Honda used an empirical formula for critical warning distance where kinematic analysis based critical brake distance was used as

$$d_w = 2.2 v_{rel} + 6.2 \quad (14)$$

$$d_{br} = \begin{cases} \tau_2 v_{rel} + \tau_1 \tau_2 a_f - 0.5 a_f \tau_1^2 & \dfrac{v_l}{a_l} \geq \tau_2 \\ \tau_2 v - 0.5 a_f (\tau_2 - \tau_1)^2 - \dfrac{v_l^2}{2 a_l} & \dfrac{v_l}{a_l} < \tau_2 \end{cases} \quad (15)$$

where $a_l$ and $a_f$ are the maximum decelerations of the leading and following vehicles, respectively. $d_{br}$ is the braking distance, $d_w$ is the warning distance, $v_f$ is the following vehicle velocity, vrel is the relative velocity, $\tau_1$ is the system delay, $\tau_2$ is the brake time [13].

*B. Intersection Collison Avoidance Systems Algorithms*

The intersection collision avoidance system algorithm developed by NHTSA-OCAR aimed at overcoming intersection collision without any direction dependence. Time to intersection was calculated for two vehicles to control whether there was any intersection collision risk. If the difference between these two time-to-intersection times was less than a predetermined parameter value, then the system would decide that there would be an eventual collision at this intersection.

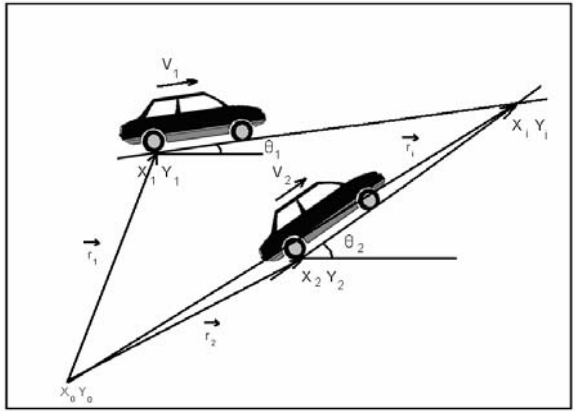

Figure 2. Intersection Collision Geometry

The time to intersection (TTX) computations are carried out by using;

$$x_i = \frac{(y_2 - y_1) - (x_2 \tan\theta_2 - x_1 \tan\theta_1)}{\tan\theta_1 - \tan\theta_2} \quad (16)$$

$$y_i = \frac{(x_2 - x_1) - (y_2 \cot\theta_2 - y_1 \cot\theta_1)}{\cot\theta_1 - \cot\theta_2} \quad (17)$$

$$TTX_1 = \frac{|\vec{r_i} - \vec{r_1}|}{|\vec{v_1}|} sign((\vec{r_i} - \vec{r_1}).\vec{v_1}) \quad (18)$$

$$TTX_2 = \frac{|\vec{r_i} - \vec{r_2}|}{|\vec{v_2}|} sign((\vec{r_i} - \vec{r_2}).\vec{v_2}) \quad (19)$$

where the variables are illustrated in Figure 2.
Warning time was determined according to driver and road characteristics as

$$TTA = t_r + \frac{\beta v}{f(\mu) g} \quad (20)$$

where r is the position vector, TTA is the time to avoidance, $t_r$ is the driver delay, β is the vehicle factor, f(μ) is the road friction factor function and f(μ)g is the effective acceleration [8].

IV.  INTER VEHICLE COMMUNICATION

Inter-Vehicle Communication (IVC) is an important part of ADAS which is open for future research applications. IVC provides infrastructure-free communication and can be used on all roads without any additional operation. The ad hoc network which is a wireless and decentralized network type supports the necessities of IVC. It is possible to apply unicast, multicast and broadcast to an ad hoc network. There are several protocols for multicast communication based on unicast. Protocols for multicast can be divided into two groups which are topology based protocols and position based protocols. Topology based protocols are also divided into parts such as proactive, reactive and hybrid protocols. Reactive protocols are on-demand protocols which obtain route information by route discovery when requested. There are two reactive protocols that dominate the literature: Dynamic Source Routing (DSR) and Ad-Hoc on-Demand Distance Vector (AODV) [17].

In DSR, shown in Figure 3, the node which wants to send a packet broadcasts a Route Request (RREQ) packet composed of destination address, source node address and a unique identification number. All nodes control whether they are the requested node or not. If they are not the requested node, then they add their identification number to the RREQ and this process continues until RREQ arrives the requested node. When RREQ arrives the requested node, the requested node sends a Route Reply (RREP) to the neighboring node which is determined according to the stored history information in RREQ and this process continues according to RREQ node information until RREP arrives the source node. It is also possible to cache in this protocol to obtain faster communication [18].

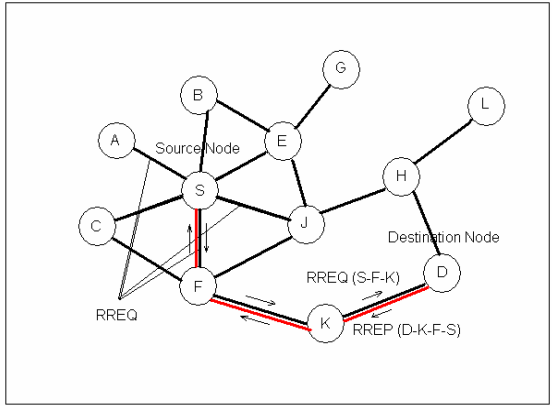

Figure 3. Dynamic Source Routing (DSR)

In AODV, RREQ does not contain node identification numbers [19]. When RREQ arrives nodes, nodes update their route tables and send these back to the source node [19]. Proactive protocols are continuous protocols that apply route discovery even when there is no communication demand [17]. In Link State Routing (LSR), each node broadcasts its status periodically, and rebroadcasts the status information which is relayed by neighbor nodes to other nodes [20]. Route tables are sent by nodes continuously with sequence numbers attached to them to avoid loops which result in redundant data communication in Destination-Sequenced Distance-Vector (DSDV) [20].

Zone Routing Protocol (ZRP) is a hybrid protocol combining proactive and reactive protocols to obtain more efficient and scalable protocol [17]. In ZRP, communication range divided into Routing Zones [21]. Communication in Routing Zones is realized by proactive protocols where communication between routing zones is realized by reactive protocols [22].

Position based protocols do not require any route discovery or route table. These protocols use position information from GPS or other information services. In position based protocols, nodes obtain the position information from Location Services which are part of the communication network. Forwarding can be realized by three different ways: Greedy Packet Forwarding, Restricted Directional Flooding and Hierarchical Routing [17].

In Greedy Packet Forwarding, shown in Figure 4, the next node is determined by different techniques. Most Forward within R (MFR) selects the node closest to the destination node to reduce hop number where Nearest with Forward Progress (NFP) selects the node closest to the source node to reduce the possibility of data collision. Another method is using the closest node which is on the straight line to the destination node. Greedy Packet Forwarding uses minimum backward progress node when a node fails to find its destination path. Restricted Directional Flooding first determines the possible area where the destination node may be in. This area is determined by location and maximum velocity information of the destination node. Next hop is selected from this area. Hierarchical Routing uses proactive routing protocol over small distances and Greedy Packet Forwarding over large distances [17].

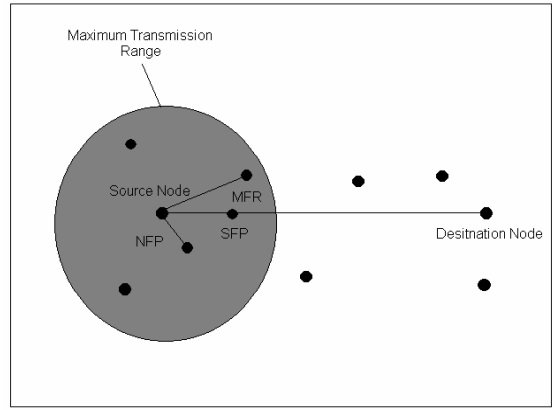

Figure 4. Greedy Packet Forwarding

Despite the fact that these multicast protocols are appropriate for Mobile Ad hoc Networks (MANET), Vehicular Ad hoc Network (VANET), significantly differs from MANET considering high mobility and the necessity of faster applications requiring more effective protocols [22]. Route discovery causes delay where route tables have possibility of storing broken routes. Broadcast which does not need any route discovery or route tables is an appropriate method for VANET.

Dedicated Omni-Purpose Inter-Vehicle Communication Linkage Protocol for Highway Automation (DOLPHIN) project used broadcast as a simple flooding mechanism for the first time [23].

However, it is expressed that simple flooding mechanism suffers from Redundant Rebroadcasts, Contention and Collision which are called Broadcast Storming Problem. Redundant Rebroadcast Problem is caused by rebroadcast desire of nodes whose neighbors have already received the data packet. Collision occurs when one node receives the same data packet from different nodes at the same time where Contention occurs when neighbor nodes rebroadcast the same data at the same time. To reduce the redundancy, collisions and contentions, five mechanisms are suggested. Probabilistic Based Algorithm suggests sending data with probability of P with random delay. Counter Based Algorithm uses threshold of C for receiving a data packet more than one time. Distance Based Algorithm uses relative distance of two nodes to calculate additional rebroadcast area provided by the next node. Location Based Algorithm uses GPS information of the nodes and convex polygon scenario to obtain additional area of broadcasting provided by the next node on the entire communication network. Cluster Based Algorithm forms a hierarchy between nodes and uses one of the above algorithms. Another Location Based Algorithm is suggested by [24]. In this algorithm, Notification and Relay distances are calculated and the desired nodes are determined according to distances with direction information which is defined by the equations

$$(\vec{X}_1 - \vec{X}_2 + \alpha \vec{V}_2) \bullet \vec{V}_2 > 0 \qquad (21)$$

$$|\vec{X}_1 - \vec{X}_2 + \alpha \vec{V}_2| < R_{relay} \qquad (22)$$

$$|\vec{X}_1 - \vec{X}_2 + \alpha \vec{V}_2| < R_{notification} \qquad (23)$$

where $\vec{X}_1$, $\vec{X}_2$, $\vec{V}_1$ and $\vec{V}_2$ are the positions and the velocities of following and leading vehicles, respectively. $R_{relay}$ is the available relay range, $R_{notification}$ is the available notification range and α is the location offset [24].

Probabilistic Based Algorithm, shown in Figure 5, is also used by [22]. To realize Probabilistic Based Algorithm, vehicle nodes are divided into parts. $SH_o$ is a set of neighbors of vehicle node o which are reached directly by the source node (one-hop neighbors), $SH^2_0$ is a set of neighbors of vehicle node o which are reached by the source with the aid of another hop (two-hop neighbors), and $M_{o,cr}$ is a set of neighbors that can be reached by the source only through by neighbor c.

Probabilistic Based Algorithm is also used in [22]. To realize Probabilistic Based Algorithm, vehicle nodes are divided into parts which are $SH_o$ is a set of one-hop neighbors of vehicle node o, $SH^0_2$ is a set of two-hop neighbors of vehicle node o, and $M_{o,cr}$ is a set of two-hop neighbors that can be reached only through one-hop neighbor c from o, r = 1, 2, 3, .....,N. By this value probability of rebroadcasting Φ is determined as follows;

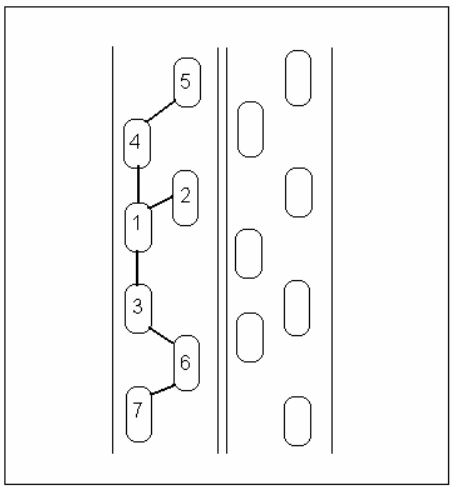

Figure 5. Probabilistic Based Broadcast (N(SHo) is 3 (vehicle 2,3 and 4), N($SH^2$o) is 2 (vehicle 5 and 6), $M_{O,4}$ is 1 (vehicle 5) and $M_{O,3}$ is 1 (vehicle 6))

Probability of rebroadcast Φ is determined as

$$\Pr_o = \begin{pmatrix} \frac{\sum_{r=1}^{N(SH_O)} N(M_{O,c_r})}{N(SH_O)} & if & \sum_{r=1}^{N(SH_O)} N(M_{O,c_r}) \leq N(SH_O) \\ 1 & \vdots & O.W. \end{pmatrix} \qquad (24)$$

$$\Pr_{o_{SH}} = \frac{N(SH_O)}{N(SH_O) + N(SH^2_O)} \qquad (25)$$

$$\Pr_{o_{SH2}} = \frac{N(SH^2_O)}{N(SH_O) + N(SH^2_O)} \qquad (26)$$

$$\Phi = \frac{\Pr_o + \Pr_{o_{SH}} + \Pr_{o_{SH2}}}{3} \qquad (27)$$

where N(SHo) is the number of one-hop neighbors, N(SH2o) is the number of two-hop neighbors and N(Mo,cr) is the number of neighbors which are reached by node c [22].

In Role Based Algorithm, routing zone is determined by the basic kinematic formula;

$$d_{br} = v \Delta t_{reaction} + \frac{v^2}{2 b_{max}} \qquad (28)$$

where v is the initial velocity of the vehicle, dbr is the braking distance, $\Delta t_{reaction}$ (reaction time) = 1 s and $b_{max}$ (maximum deceleration) = 4.4 m/s$^2$ [26]. After determining the routing zone to avoid redundant rebroadcasts, the delay time is determined to avoid collisions and contentions using

$$WT(d) = -\frac{Max(WT)}{Range} \overset{*}{d} + Max(WT) \qquad (29)$$

$$\overset{*}{d} = \min(d, Range) \qquad (30)$$

where d is the distance, WT(d) is the waiting time for distance d, Max(WT) is the maximum waiting time and Range is the transmission range [26].

Hidden nodes are another important problem of broadcasting. Reference [27] suggests a new mechanism which solves this problem in urban areas. The suggested algorithm is composed of two parts which are directional and intersectional broadcast. In directional broadcast, the road is divided into segments. Furthest node is selected to be the next relay node. If road segments have more than one node, then it is divided into sub segments until the obtained segment size is appropriate. Furthest node sends and acknowledges confirmation of successful reception. When vehicles arrive at an intersection, then routers at the intersection operate to send packets in each road segment. These packets are transmitted through repeaters directionally [27].

In multicasts, collisions are avoided by Medium Access Control (MAC) layers. Carrier Sense Multiple Access with Collision Avoidance (CSMA/CA) is used in many wireless communication systems. Node controls the available channel to send data packets to desired nodes. If the medium is idle, it sends the packet, if it is busy; it waits for DCF Inter-Frame Space (DIFS) interval. The Hidden Terminal problem is solved by Request-To-Send/Clear-To-Send (RTS/CTS) handshake in CSMA/CA. After reaching the desired node with RTS/CTS, the desired node sends Acknowledges (ACKs) to the source node [28].

MAC layers can be applied to the broadcasts. A reliable MAC layer which can also detect data errors is suggested by reference [29]. Broadcast Medium Window (BMW) has Request-To-Send/Clear-To-Send (RTS/CTS) handshake with sequence numbers. Each node has RECEIVER and SEND buffer which contain received and sent data. Node controls the sequence number of the RTS with the desired sequence number. If there is missing data, it sends the CTS with the sequence number of the desired one to the source node. Intermediate nodes which hear CTS and have desired data can send this data. There is also a sign for end of data. If node does not receive end of data, it sends CTS after waiting a random amount of time [29].

Reference [30] suggests an adaptive data layer for efficient Inter-Vehicle Communication. Data transmittal in communication area is classified into groups such as Class 1: Position, Velocity and Emergency Message, Class 2: Road Condition, Sudden Braking and Information about Obstacles; Class 3: Turn Indication, Acceleration, Steering Angle, and Sudden Braking. The data layer is divided into segments with desired ratios of data classes. Vehicles are also classified into groups according to the data queue. Data transmits through vehicles with reducing amount. By this technique redundancy is avoided. It is also possible to apply multimedia transmissions to this IP similar structure [30]. Apart from this application, there are also Inter-Vehicle Communications which are combined with the road infrastructure. Fleetnet is one of them and aims at combining an ad hoc network with wireless internet applications [31]. It is also possible to communicate using spread spectrum method. In this network, every vehicle uses the PN coded pulse radar system to obtain other vehicles' position information and magnetic markers are used to obtain high accuracy [32]-[33].

Standardization is a very important problem of the inter-vehicle communication systems. Vehicles have to talk same language to obtain viable IVC systems and it is only realized by standard communication protocols. Despite the fact that all research efforts provide visible progressions on overcoming broadcast storm problem, there is no protocol which is accepted by the wide range of researchers who study this subject.

V. CONCULUSION

This paper aimed at giving information on recent developments on CA/CW System and Inter-Vehicle Communication Systems. The importance of CA/CW system was presented. It was seen that CA/CW systems were important ADAS, considering the number of fatal accidents that can be prevented. Economic advantages of CA/CW System were observed from research results of these systems, such as road capacity rises and fuel consumption reductions. Status of finished and on-going projects all over the world was also presented. Information on Inter-Vehicle Communication Systems was also presented.

The wide research area of CA/CW systems is waiting for researchers who want to work on this important problem. The most important barrier that must be overcome is the human factor. Developed systems must be compatible with the drivers. The second barrier is legal. Laws for controlling these technologies must be arranged. The problems of Inter-Vehicle Communication are also an important barrier in front of the progress of CA/CW Systems. With the new protocols solving broadcasting problems and providing efficient and scalable communication, it will be possible to obtain fully communicating vehicles. Improvements in sensor technologies are also required to develop more advanced systems.

ACKNOWLEDGMENT

The authors acknowledge the support of the European Union Framework Programme 6 through the AUTOCOM SSA project (INCO Project No: 16426) in this paper.